\documentclass{llncs}

\usepackage[english]{babel}
\usepackage[utf8]{inputenc}
\usepackage{amsmath}
\usepackage{graphicx}
\usepackage{url}
\usepackage{amssymb}
\usepackage{booktabs}

\title{Pruning Techniques for Mixed Ensembles of Genetic Programming Models}

\author{Mauro Castelli\inst{1} \and 
Ivo Gon\c{c}alves\inst{2,3} \and Luca Manzoni\inst{4} \and Leonardo Vanneschi\inst{1}}

\institute{NOVA IMS, Universidade Nova de Lisboa, 1070-312 Lisboa, Portugal\\
  \and INESC Coimbra, DEEC, University of Coimbra, Pólo 2, 3030-290 Coimbra, Portugal\\
  \and CISUC, Department of Informatics Engineering,\\
  University of Coimbra, 3030-290 Coimbra, Portugal
  \and Dipartimento di Informatica, Sistemistica e Comunicazione,\\
  Universit\`a degli Studi di Milano-Bicocca, 20126 Milano, Italy\\
  \email{mcastelli@novaims.unl.pt, icpg@dei.uc.pt,\\
    luca.manzoni@disco.unimib.it, lvanneschi@novaims.unl.pt}}

\begin{document}

\maketitle
 
%
%

\begin{abstract}

The objective of this paper is to define an effective strategy for building an ensemble of Genetic Programming (GP) models. Ensemble methods are widely used in machine learning due to their features: they average out biases, they reduce the variance and they usually generalize better than single models. Despite these advantages, building ensemble of GP models is not a well-developed topic in the evolutionary computation community. To fill this gap, we propose a strategy that blends individuals produced by standard syntax-based GP and individuals produced by geometric semantic genetic programming, one of the newest semantics-based method developed in GP. In fact, recent literature showed that combining syntax and semantics could improve the generalization ability of a GP model. Additionally, to improve the diversity of the GP models used to build up the ensemble, we propose different pruning criteria that are based on correlation and entropy, a commonly used measure in information theory. Experimental results, obtained over different complex problems, suggest that the pruning criteria based on correlation and entropy could be effective in improving the generalization ability of the ensemble model and in reducing the computational burden required to build it.
\end{abstract}

%
%

\section{Introduction}
\label{s1}

In the last few years, effort was dedicated to the definition and analysis of methods to exploit semantic awareness in GP~\cite{survey}, where the term semantics generally refers to the behavior of a program when executed on a set of training cases. Semantics-based methods provide a new conceptual view on GP and were successfully used to solve complex problems over different domains~\cite{Castelli201537,castelli2013efficient}. However, standard syntax-based GP is also capable of obtaining competitive results in different fields~\cite{Yoo:2017:HCG:3092955.3078840,Picek:2017:ESB:3067695.3076084}. In both cases, the application of a GP algorithm is generally performed with the objective of obtaining a final model able to fit the training data as best as possible. So far, a little research effort was dedicated to the construction of ensemble models based on GP~\cite{Keijzer2000} and this is somehow surprisingly considering the vast amount of literature where the advantages of ensemble methods are reported~\cite{Dietterich2000}. To formally define an ensemble model let us refer to a standard symbolic regression problem since this is the kind of application addressed in this study. In symbolic regression, the goal is to search for the symbolic expression $f(\vec{x})$ that best fits a particular training set $T = \left\lbrace(\vec{x_1}, y_1), \ldots, (\vec{x_n}, y_n)\right\rbrace$ of $n$ input/output pairs with $\vec{x_i} \in \mathbb{R}^n$ and $y_i \in \mathbb{R}$.
An ensemble of regression models is a set of symbolic expressions whose individual predictions are combined (typically by considering their median or a weighted sum) to predict the target $y_i$. Interestingly, ensembles are often much more accurate than the individual predictors that make them up~\cite{Dietterich2000}. The main reasons can be summarized as follows: (1) a learning algorithm performs a search in a space $H$ of hypotheses to identify the best hypothesis in the space. When the number of training data is too small compared to the size of the hypothesis space, the learning algorithm can return different hypotheses in $H$ and all of them present an error on the training instances. In such a situation, an ensemble can reduce the bias toward a particular hypothesis simply by considering different hypothesis and returning a prediction based on the predictions of \textit{all} the hypothesis that made up the ensemble. (2) In the large part of the problems addressed by using machine learning techniques, the target function $f$ cannot be represented by any of the hypotheses in $H$. In this case, an ensemble can expand the space of representable hypothesis by considering weighted sums of hypotheses drawn from $H$

One of the main reasons for the poor attention dedicated to ensembles in the GP literature may be related to the fact that, since its inception, GP was considered a time-consuming process due to the computational complexity required by the evaluation of the fitness function. While this issue is not critical for semantics-based GP~\cite{imp,goncalves2017stopping} and despite the availability of effective hardware that nowadays allows to perform fast parallel computations, the construction of ensemble models based on GP has not received the attention it deserves.

To answer this call, this study presents a method to effectively build ensembles of GP models. The method is designed in such a way to overcome the main limitations of the typical approach used for building an ensemble model, where different parallel GP populations are evolved and the final ensemble will be composed of the best models returned by each one of the these populations. In fact, while this approach is the predominant one across the literature~\cite{Dietterich2000,polikar2012ensemble}, it is important to consider several issues that have a negative impact on the ensemble model developed with such an approach. One of the most important issues is the one related with generalization. Generalization refers to the ability of a model to perform well on unseen examples. This is a critical aspect of a model and the interest in studying generalization in GP has been recently increasing~\cite{goncalves2017stopping,GoncalvesPhdThesis,Chen2016,gonccalves2015generalization,Kommenda2014,Goncalves2013,Goncalves2012,Goncalves2011,castelli2011quantitative}. Its importance is related to the fact that, typically, the final user of a model wants to obtain satisfactory performance on new instances of the problem at hand, while the performance on the training cases is generally irrelevant. For this reason, this study takes into account different approaches that we expect to be beneficial in increasing the generalization ability of the ensemble model. The first idea comes from recent literature~\cite{vanneschi2017initialization,vanneschi2017parallel} where authors demonstrated that a blend of individuals created with standard syntax-based GP (STGP) and Geometric Semantic Genetic Programming (GSGP) results in a model with a better generalization ability with respect to the use of only one kind of solutions. With this in mind, to build the ensemble of GP models we run in parallel different GP populations, where some of them are evolved using STGP and others are evolved using GSGP (as done in~\cite{vanneschi2017parallel}). One of the hypotheses of this study is that a blend of STGP and GSGP is beneficial (in terms of generalization) also in building an ensemble of GP models.

The second idea is related to the fact that running all the populations for a given number of generations may result in an unwanted behavior, where all the evolved individuals are semantically similar (i.e., they produce approximately the same outputs for all the different training cases). In such a situation, there would be a little advantage in using an ensemble with respect to the usage of a single model. In fact, as reported in~\cite{hansen1990neural} an ensemble should be composed of models that are accurate and \textit{diverse}. In this context, two models are said to be diverse if they make different errors on new data. It is only in this way that it would be possible to achieve a better generalization ability with respect to the one achievable by considering a single model. To take into account this relevant aspect, the strategy presented in this work will evolve different parallel populations while some of them are removed by using different similarity-based pruning criteria. The criteria that are considered in this study are based on the level of entropy and correlation of the best models available in the populations. The use of these pruning criteria should guarantee some sort of diversity among the models used to build up the ensemble. 

The paper is structured as follows: Section~\ref{s2} briefly presents some previous studies related to the definition of ensemble models. Section~\ref{s3} presents the strategy developed in this study to build an ensemble of GP models and the similarity-based criteria we defined. Section~\ref{s4} contains the experimental study, including a presentation of the used test problems and of the experimental settings and a discussion of the obtained results. Finally, Section~\ref{s5} concludes the paper and suggests possible avenues for future work.

%
%

\section{Related Work}
\label{s2}

This section presents recent contributions related GSGP as well as existing studies involving ensemble models and GP.

The use of ensemble models and GP presents only a few contributions in the literature. One of the first studies dates back to 2000 when a study about the decomposition of regression error into bias and variance terms to provide insight into the generalization capability of modelling methods was proposed~\cite{Keijzer2000}. After an introduction to bias/variance decomposition of mean squared error, authors showed how ensemble methods such as bagging~\cite{breiman1996bagging} and boosting~\cite{freund1996experiments} can reduce the generalization error in GP. Bagging and boosting were considered in the context of ensemble models for GP in~\cite{iba1999bagging}. In their work, authors presented an extension of GP by means of resampling techniques. By considering bagging and boosting, they manipulated the training data in order to improve the learning algorithm. In their work they extended GP by dividing a whole population into a set of sub-populations, each of which is evolved by using the bagging and boosting methods. Best individuals of each sub-population participate in voting to give a prediction on the unseen data. The performance of their approach was discussed and authors also showed the beneficial effect of the proposed technique in reducing bloat with respect to the standard GP algorithm. A study related to the suitability of EC techniques in building ensemble models for classification tasks was presented in~\cite{gagne2007ensemble}, where authors presented the so-called Evolutionary Ensemble Learning (EEL) approach. The objective of the study was twofold: on one side they defined a new fitness function inspired by co-evolution to enforce the classifier diversity. Additionally, a new selection criterion based on the classification margin is proposed. The new selection criterion is used to extract the classifier ensemble from the final population or incrementally along the evolution. In the experimental phase, they showed the suitability of their approach when compared to a single-hypothesis evolutionary learning process.

Besides the aforementioned theoretical studies, ensemble models and GP were used to solve complex real-world problems, mainly related to classification tasks. In~\cite{ZHANG200485} authors demonstrated the suitability of GP as a base classifier algorithm in building ensembles for large-scale data classification. In particular, they showed that an ensemble of GP individuals is able to significantly outperform its counterparts built upon base classifiers that were trained with decision tree and logistic regression. Authors also claimed that the superiority of GP ensemble is partly attributed to the higher diversity, both in terms of the functional form as well as with respect to the variables defining the models, among the base classifiers upon which it was built on. In the same context of large-scale data classification, an extension of cellular genetic programming for data classification (CGPC) to induce an ensemble of predictors is presented in~\cite{folino2006gp}. In their work authors developed two algorithms based on bagging and boosting and compared their performance with the one of CGPC. Results showed that the proposed approaches are able to deal with large datasets that do not fit in main memory, also producing better classification accuracy with respect to standard CGPC. The same authors proposed the use of GP ensemble for distributed intrusion detection systems~\cite{folino2005gp}. The algorithm runs on a distributed hybrid multi-island model-based environment to monitor security-related activity within a network. Experiments showed the validity of the approach when compared to standard techniques for the task at hand. Other applications of ensemble methods to GP includes the use of querying-by-committee methods~\cite{isele2013active,bartoli2017active} and of a divide-and-conquer strategy, in which ax solution need to work well only on a subset of the entire training set~\cite{pappa2009evolving,bartoli2015learning}

With respect to ensembles of regression models, a quite recent contribution was proposed in~\cite{veeramachaneni2013learning}. The idea explored by the authors was to generate several regression models by concurrently executing multiple independent instances of a GP and, subsequently to analyze several strategies for fusing predictions from the multiple regression models. The study considered only small datasets due to memory constraints, but authors were able to draw interesting conclusions about the suitability of their approach in producing accurate predictions. Our study will differ from the one described in~\cite{veeramachaneni2013learning} in several ways: we do not put any constraint on the size of the datasets, we will consider models produced by different GP algorithms (blend of STGP and GSGP) and we define and use different similarity-based criteria that, by taking into account the information related to all the populations evolved, aim at improving the generalization ability of the final ensemble as well as reducing the computational effort. Hence, in the experiments described in this contribution and as explained in Section~\ref{s3}, the populations evolved are not independent of each other.

As reported in Section~\ref{s1}, one idea exploited in this study is to build an ensemble that consists of a blend of individuals produced by STGP and GSGP. GSGP is one of the newest methods to directly include semantic awareness in the search process~\cite{PaperMoraglio}. The interested reader is referred to~\cite{survey} for a description of the concept of semantics and its uses in GP, while~\cite{PaperMoraglio,vanneschi:2013:EuroGP} present the semantic operators for GP used in this paper. Despite the plethora of studies investigating the role of semantics, this is still a hot topic in the field of GP. Particularly interesting with respect to our study is the work proposed in~\cite{vanneschi2017initialization}. In their work, authors defined a simple yet effective algorithm for the initialization of a GP population inspired by the biological phenomenon of demes despeciation (i.e. the combination of demes of previously distinct species into a new population). In synthesis, the initial population of GP is created using the best individuals of a set of separate subpopulations, or demes, some of which run STGP and the others GSGP for few generations. Experimental results showed that this initialization technique outperforms GP with the traditional ramped half-and-half algorithm on six complex symbolic regression applications. Even more interesting, by using the proposed initialization technique, the GP process produces individuals with a better generalization ability than the ones obtained by initializing the population with the traditional ramped half-and-half algorithm. Hence, to construct the GP-based ensemble we build upon this idea and we expect to obtain a final ensemble with a better generalization ability with respect to its counterparts, where only STGP or GSGP are considered.

\section{Method}
\label{s3}

This section describes the proposed system for building GP-based ensemble models. The main idea is to provide a pruning method to reduce the number of populations in an ensemble when they are exploring similar regions of the search space and, in some sense, they are possibly wasting computational effort to perform the same work two times. Therefore, we need a measure of similarity among solutions that allows the pruning procedure to take place. After each generation, all the  best solutions for all the populations that are part of the ensemble are pairwise compared and, if two of them are deemed too similar, the worst performing one (in terms of fitness) is removed and the one remaining is now weighted more when calculating the semantics of the ensemble (that is a weighted sum of the semantics of the best solutions).

Formally, let $P_1, \ldots, P_n$ be $n$ populations, and let $I_1, \ldots, I_n$ be their best individuals each one having semantics $s(I_i)$. Each population has associated a weight $w_i$ (all the weights are equal to $1$ after initialization) and the semantics of the entire ensemble is given by $\frac{1}{n} \sum_{i=1}^n w_i s(I_i)$. After each generation, a subroutine $D(I_i,I_j)$ that calculates the similarity between $I_i$ and $I_j$ is called for each of the best individuals of the populations. Since many similarity measures returns a real value measuring how similar the two individuals are, we obtain a Boolean answer by comparing the similarity measure with a threshold. After that, if True is returned (i.e., $I_i$ and $I_j$ are similar), the fitness $f(I_i)$ and $f(I_j)$ are compared and the population corresponding to the worst fitness is removed. For example, if $f(I_j)$ is the worst fitness, then $P_j$ is removed from the ensemble and $P_i$ will increase its weight from $w_i$ to $w_i + w_j$. The main aspect that governs this process is the subroutine $D(I_i,I_j)$, and we are going to describe four different implementations of it based on two different notions of semantic similarity: entropy and correlation.

\subsection{Correlation-based Similarity}

The idea of a correlation-based similarity is to consider two semantics as similar if the correlation among them is above a certain threshold. Let $s(I_i)$ and $s(I_j)$ be two semantics, i.e., two semantic vectors, and let $\rho_{i,j}$ be the Pearson correlation coefficient among them. Its value varies between $-1$ (negative correlation) to $1$ (positive correlation). If the correlation is higher than $0.5$ we consider the two individual similar enough. This threshold was selected after a preliminary tuning phase, where different values were tested across the benchmarks considered.

A variation of this method introduces a probability of being considered similar enough when the correlation coefficient goes above the threshold. The probability for $s(I_i)$ and $s(I_j)$ to be considered equal is set to $\rho_{i,j}$ (i.e., if $D(s(I_i), s(I_j))$ is greater than the threshold value, then true is returned with probability $\rho_{i,j}$). This will reduce the expected number of populations that will be removed from the ensemble while still assuring that very similar ones (i.e., with the correlation coefficient near one) will almost surely be removed. 

\subsection{Entropy-based Similarity}

\newcommand{\Inf}{\mathcal{I}}

The entropy-based similarity is based on the idea that, if two semantics are similar, it should be possible to infer the outputs of one based on the other, This is possible even if the relation is more complex (non-linear) than the one that can be captured by using the linear correlation coefficient. Given two semantics $s(I_i)$ and $s(I_j)$, we will denote their mutual entropy by $H(i,j)$ and their mutual information by $\Inf(i,j)$. Notice that to compute these values we need to provide discrete data, which is not the case for the semantics of GP individuals under regression problems. Therefore, we discretize them using $\sqrt{n}$ equally-sized bins, with $n$ the length of the semantic vectors, and counting the number of elements that are present in each bin.

The similarity measure is based on the \emph{variation of information}, that is, the metric $d(i,j) = H(i,j) - \Inf(i,j)$. To normalize it between $0$ and $1$ we use $D(i,j) = \frac{d(i,j)}{H(i,j)}$. This distance will be close to $1$ if the two semantics are dissimilar and close to $0$ otherwise. As a threshold the value $0.5$ was chosen. Thus, two semantics with distance lower than $0.5$ are considered similar enough. Also in this case, the threshold was selected after a preliminary tuning phase, where different values were tested.

As with the correlation-based similarity, it is possible to introduce a probabilistic variation, in which a population is selected for deletion with probability $1-D(i,j)$ when $D(i,j) < 0.5$.

\section{Experimental Settings and Results}
\label{s4}

Five datasets were considered for testing the performance of the GP-based ensemble. These datasets were already considered in previous GP studies. Hence, we summarize their main properties in Table~\ref{tab1} by reporting the number of independent variables, the name of the problem and a reference where readers may find detailed descriptions of the datasets. In these experimental phase, benchmarks with real-world data were chosen to establish the suitability of the proposed methods in a real-world setting. 

\begin{table}[!ht]
  \caption{\label{tab1} Description of the test problems. For each dataset, the number of features (independent variables) and the number of instances (observations) are reported.}
  \centering
  \begin{tabular}{rrr}
    \toprule
    Dataset & \# Features & \# Instances \\
    \midrule
    Airfoil~\cite{airfoil} & 5     & 1502 \\
    Concrete~\cite{Castelli20136856} & 8     & 1029 \\
    Protein Plasma Binding Level~(PPB)~\cite{castelli2016parameter} & 626   & 131 \\
    Slump~\cite{slump} & 9     & 102 \\
    Yacht~\cite{ortigosa2007neural} & 6     & 307 \\
    \bottomrule
  \end{tabular}
\end{table}%

The objectives of the experiment are the following:
\begin{itemize}
  \item to show that the pruning criteria can effectively remove some of the existing populations;
  \item that the removal of populations is performed in a way that does not deteriorate performance;
  \item that the pruning criteria perform better than simply randomly removing populations or selecting a smaller ensemble size.
\end{itemize}
Here, the notion of being better can be interpreted in two ways: either a performance improvement with respect to fitness or comparable fitness values obtained using a smaller number of populations. That is, the proposed pruning criteria either improve the results or reduce the computational burden.

The ensembles are composed, at the beginning, of $20$ populations, half of them will evolve by using STGP and the other half by using GSGP. To generate the training set, for each run a global training set consisting of $70\%$ of the problem instances was selected. The remaining $30\%$ was used as the test set. From the training set, each population in the ensemble was provided with a local training set consisting of the same number of observations but obtained by randomly sampling with replacement from the global training set. This was performed for each of the $30$ runs.

The following methods were compared:
\begin{itemize}
  \item \textbf{Standard}. No deletion of populations.
  \item \textbf{Random}. Each population has a probability of $0.001$ of being removed at each generation. At least one population will always remain.
  \item \textbf{Half}. At the start, half of the populations are selected (one half of them using STGP and the other half using GSGP) and no further deletion occurs.
  \item \textbf{Correlation}. The correlation-based similarity is used in the pruning.
  \item \textbf{Correlation-prob}. The probabilistic variation of the correlation-based similarity is used in the pruning.
  \item \textbf{Entropy}. The entropy-based similarity is used in the pruning.
  \item \textbf{Entropy-prob}. The probabilistic variation of the entropy-based similarity is used in the pruning.
\end{itemize}

The general parameters of the system are summarized in Table~\ref{tab2}. The values of the parameters were selected taking into account the values already used in existing literature, where the datasets considered in this study were already used as benchmarks. For the sake of brevity, we report only the results obtained on the test set and avoid reporting the global training set results. To test the statistical significance of the results, we have used the single tailed Mann-Whitney U-test with the alternative hypothesis that the first series of fitness values is lower (i.e., better) than the second series. As a threshold for the p-value we selected $\alpha=0.05$. 
  
\begin{table}[!tb] \centering
  \caption{\label{tab2} Parameters used in the experiments}
  \begin{tabular}{ll} \textbf{Parameter} & \textbf{Value} \\
    \hline \hline \noalign{\smallskip} Runs & 30\\
    Generations                             & 1000 \\
    Population size                         & 200 \\
    Training - Testing division             & 70\% - 30\% \\
    Fitness                                 & Root Mean Squared Error \\
    Crossover probability                   & 0.6 \\
    Mutation probability                    & 0.3 \\
    Tree initialization                     & Ramped Half-and-Half, \\
    ~                                       & maximum depth 6 \\
    Function set                            & +, -, *, and protected / \\
    Terminal set                            & Input variables, \\
    ~                                       & no constants \\
    Parent selection                        & Tournament of size 4 \\
    Elitism                                 & Best individual always survives \\
    Maximum tree depth                      & None \\
    Ensemble size                           & 20 (10 STGP and 10 STGP) \\
    \noalign{\smallskip} \hline \hline
  \end{tabular}
\end{table}

\begin{figure}
  \centering
  \includegraphics[width=0.7\textwidth]{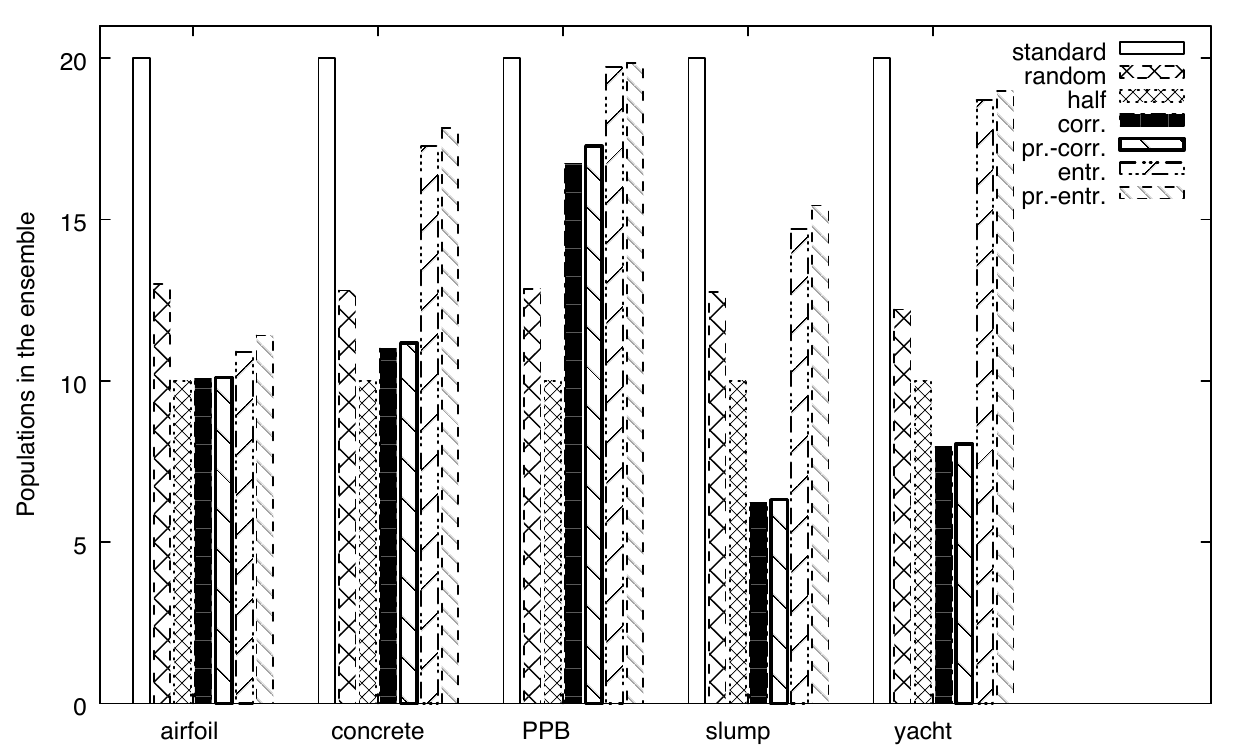}
  \caption{\label{fig:ensemble-size} The average number of populations in the ensemble for the considered methods}
\end{figure}

\begin{figure}
  \centering
  \includegraphics[width=0.68\textwidth]{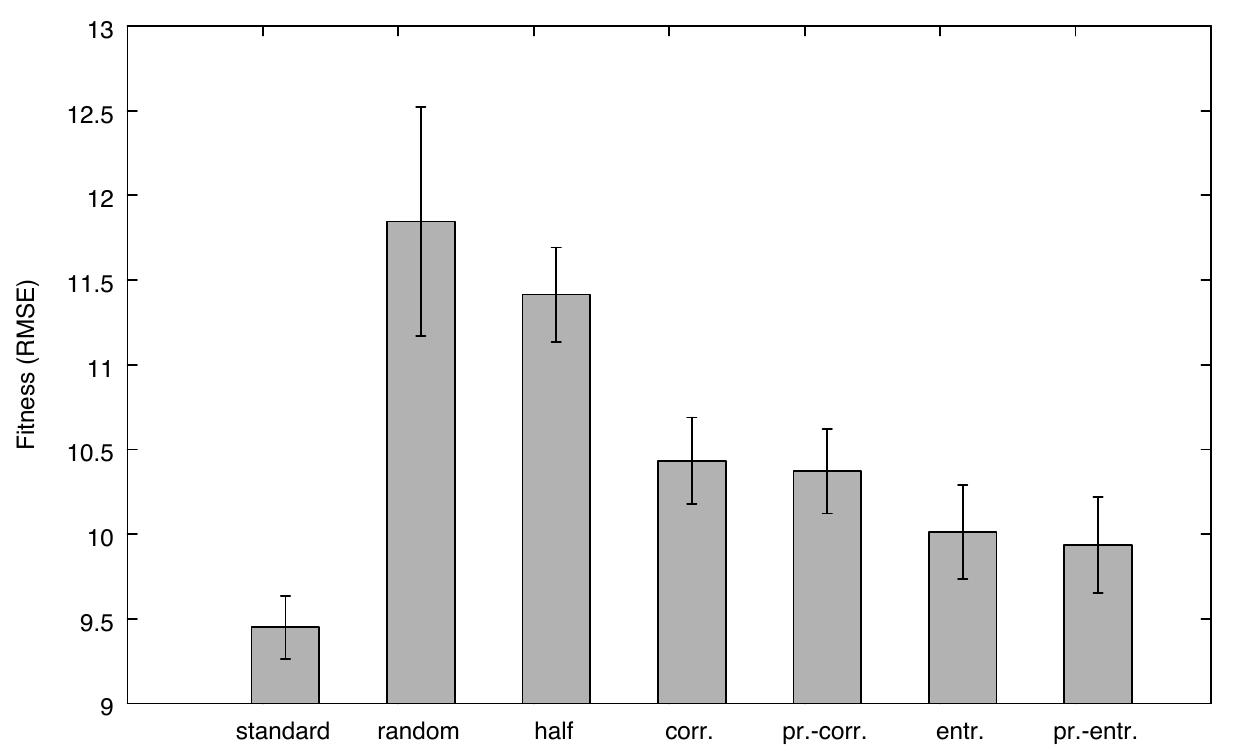}
  \caption{\label{fig:airfoil-test} The average fitness at the last generation in the airfoil dataset. The error bars are one standard deviation in length.}
\end{figure}

\begin{figure}
  \centering
  \includegraphics[width=0.68\textwidth]{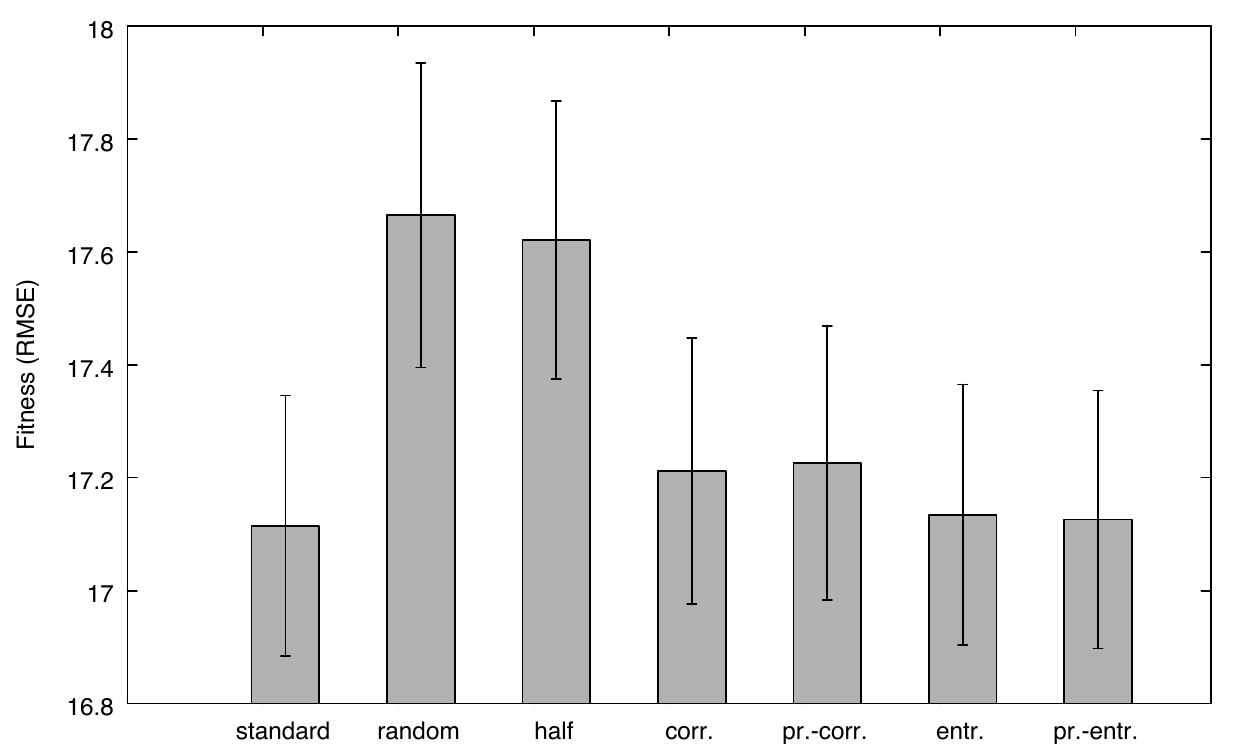}
  \caption{\label{fig:concrete-test} The average fitness at the last generation in the concrete dataset. The error bars are one standard deviation in length.}
\end{figure}

\begin{figure}
  \centering
  \includegraphics[width=0.68\textwidth]{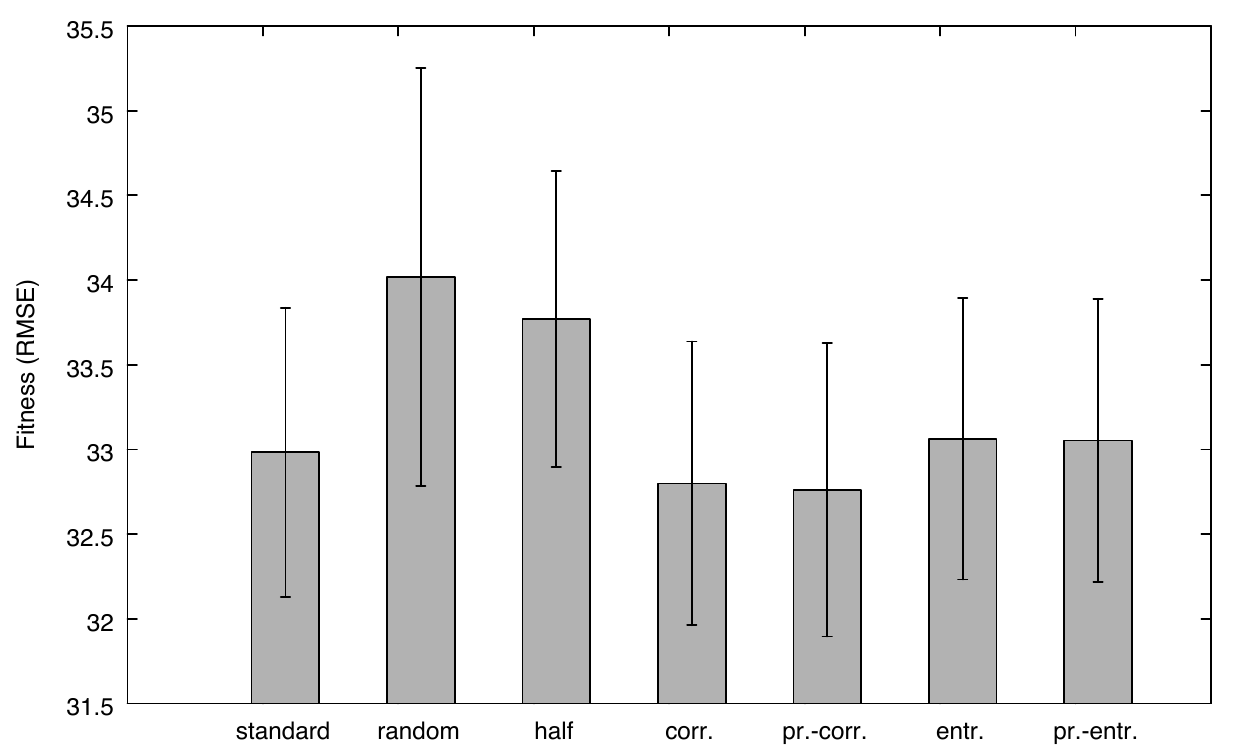}
  \caption{\label{fig:ppb-test} The average fitness at the last generation in the PPB dataset. The error bars are one standard deviation in length.}
\end{figure}

\begin{figure}
  \centering
  \includegraphics[width=0.68\textwidth]{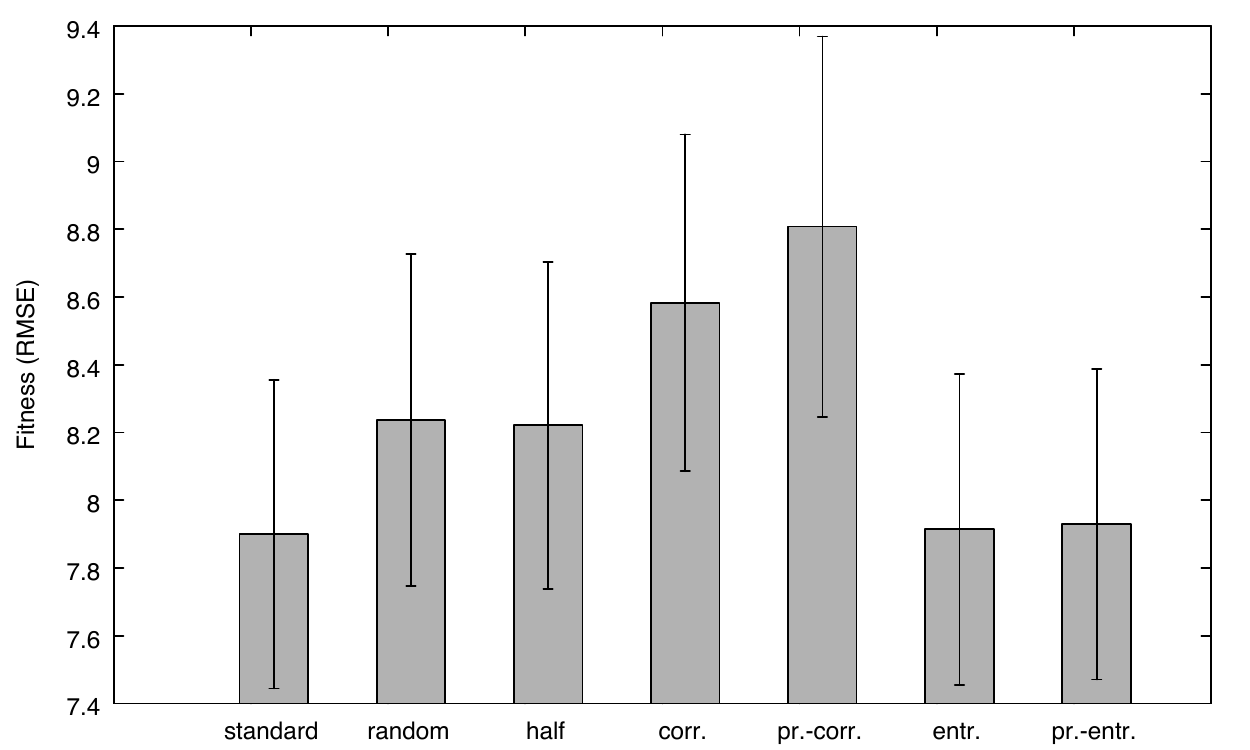}
  \caption{\label{fig:slump-test} The average fitness at the last generation in the slump dataset. The error bars are one standard deviation in length.}
\end{figure}

\begin{figure}
  \centering
  \includegraphics[width=0.68\textwidth]{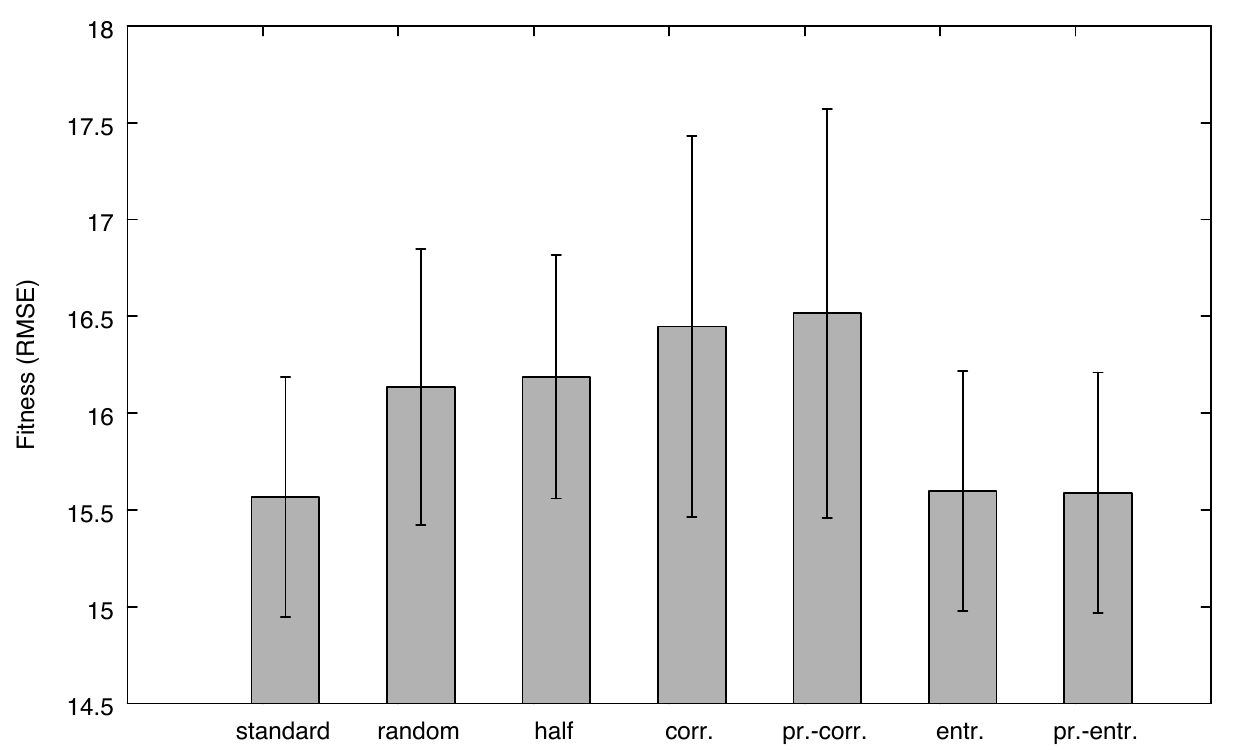}
  \caption{\label{fig:yacht-test} The average fitness at the last generation in the yacht dataset. The error bars are one standard deviation in length.}
\end{figure}

Tables~\ref{tab:airfoil}--\ref{tab:yacht} show the results of the statistical tests, where the entry in row $i$ and column $j$ is the p-value of the Mann-Whitney U-test where the technique in the $i$-th row is compared against the one in the $j$-th column. A value less then $0.05$ (highlighted in bold) indicates that we have accepted the alternative hypothesis and, thus, the $i$-th method produces lower (i.e., better) fitness values than the $j$-th one on the test data. Figures~\ref{fig:airfoil-test}--\ref{fig:yacht-test} show, for the test problems considered, the fitness on the test set for the different methods. The average size of the ensemble across all generations and all runs is shown in Figure~\ref{fig:ensemble-size}.

The results on the airfoil dataset (Figure~\ref{fig:airfoil-test} and Table~\ref{tab:airfoil}) show that the standard ensemble method is the best performer. All the four proposed methods, however, perform better than simply halving or randomly removing populations from the ensemble. Figure~\ref{fig:ensemble-size} shows that those results were obtained while using, on average, about half of the populations than the standard method.

For the concrete dataset (Figure~\ref{fig:concrete-test} and Table~\ref{tab:concrete}), the standard method and all the four proposed methods perform in a similar way. Using only half of the populations or randomly removing them produces worse results. In this case, the two correlation-based methods also use about half of the populations employed by the standard method.

For the PPB dataset (Figure~\ref{fig:ppb-test} and Table~\ref{tab:ppb}) the correlation-based methods return the lowest (i.e., better) fitness but the difference is not statistically significant when compared to the standard and entropy-based methods. The other two methods are the worst performers. The best results are obtained by using quite a high number of populations for all the proposed methods. This might indicate that the PPB problem is well-suited to be solved with ensemble techniques and that additional populations help in producing better results.

The slump dataset (Figure~\ref{fig:slump-test} and Table~\ref{tab:slump}) has the standard and the entropy-based methods as the best performers. The worst performers are the correlation-based methods which, in this case, also employ a very low number of populations.

The yacht dataset (Figure~\ref{fig:yacht-test} and Table~\ref{tab:yacht}) is interesting since it provides an example in which two of the proposed methods, namely the correlation-based ones, perform the worse, while the entropy-based remains on-par with the standard method at the price of removing only a few populations.

To conclude this section, it is important to discuss the competitive advantage in considering a blend of STGP and GSGP. To evaluate this aspect, we compared the performance of the ensemble model built considering a blend of STGP and GSGP without pruning, against the ones obtained by considering only STGP and only GSGP. 
The results achieved on the five benchmarks (not reported here due to the page limit) show that no statistically significant difference exists with respect to the performance of the considered ensemble models on unseen instances. While this result seems to contradict recent studies (e.g., \cite{vanneschi2017initialization}), a deeper analysis is needed. In particular, it would be interesting to study how the number of models in the final ensemble affects the generalization ability of the three systems (only STGP, only GSGP, blend of STGP and GSGP) and to analyze the impact of the pruning techniques on the generalization error.
Despite the fact that no statistically significant differences can be noticed, using a blend of STGP and GSGP might still be important and non detrimental: results show that the diversity of the model is generally greater than the one observable by considering a pool of individuals obtained with only GSGP (or STGP) and, additionally, the pruning criteria can determine what are the populations and models that should evolve. Hence, it might be advisable to use both GP systems to evolve the weak models and to let the pruning criteria to select the one needed to create a competitive ensemble model. As for the advantage in term of runtime and number of fitness evaluations, it is important to notice that a pruning method that, for example, reduces on average the number of distinct populations used by one third, also reduces the number of fitness evaluations by one third and, since the computation of the similarity is usually not the most computationally intensive part of the GP algorithm, a similar improvement is reflected in the reduction of the runtime.

\begin{table}
  \caption{\label{tab:yacht}\label{tab:slump}\label{tab:ppb}\label{tab:concrete}\label{tab:airfoil} p-values for the statistical tests on the airfoil, concrete, PPB, slump, and yacht datasets (from top to bottom).}
  \centering
  \begin{tabular}{l|p{1.3cm}p{1.3cm}p{1.3cm}p{1.3cm}p{1.3cm}p{1.3cm}p{1.3cm}}
    \textbf{airfoil}
    & standard & random & half & corr. & pr.-corr. & entr. & pr.-entr. \\
    \hline
    standard         & &\textbf{0.000} &\textbf{0.000} &\textbf{0.000} &\textbf{0.000} &\textbf{0.0000} & \textbf{0.001} \\
    random           & 1.000 & &0.886 &1.000 &1.000 &1.000 &1.000 \\
    half             & 1.000 &0.117 & &1.000 &1.000 &1.000 &1.000 \\
    correlation      & 1.000 &\textbf{0.000} &\textbf{0.000} & &0.677 &0.996 &0.999 \\
    prob-correlation & 1.000 &\textbf{0.000} &\textbf{0.000} &0.329 & &0.990 &0.998 \\
    entropy          & 1.000 &\textbf{0.000} &\textbf{0.000} &\textbf{0.004} &0.010 & &0.735 \\
    prob-entropy     & 1.000 &\textbf{0.000} &\textbf{0.000} &\textbf{0.001} &\textbf{0.002} &0.270 & \\
    \hline
    \textbf{concrete}
    & standard & random & half & corr. & pr.-corr. & entr. & pr.-entr. \\
    \hline
    standard         & &\textbf{0.000} &\textbf{0.000} &0.190 &0.156 &0.432 &0.476 \\
    random           & 1.000 & &0.720 &0.999 &0.999 &1.000 &1.000 \\
    half             & 1.000 &0.285 & &0.999 &0.999 &1.000 &1.000 \\
    correlation      & 0.814 &\textbf{0.001} &\textbf{0.001} & &0.456 &0.749 &0.777 \\
    prob-correlation & 0.848 &\textbf{0.001} &\textbf{0.002} &0.550 & &0.802 &0.814 \\
    entropy          & 0.573 &\textbf{0.000} &\textbf{0.000} &0.255 &0.202 & &0.527 \\
    prob-entropy     & 0.529 &\textbf{0.000} &\textbf{0.000} &0.228 &0.190 &0.479 & \\
    \hline
    \textbf{PPB}
    & standard & random & half & corr. & pr.-corr. & entr. & pr.-entr. \\
    \hline
    standard         & &0.050 &\textbf{0.034} &0.685 &0.695 &0.418 &0.424 \\
    random           & 0.952 & &0.591 &0.971 &0.974 &0.927 &0.931 \\
    half             & 0.967 &0.415 & &0.984 &0.984 &0.953 &0.956 \\
    correlation      & 0.321 &\textbf{0.030} &\textbf{0.016} & &0.529 &0.241 &0.260 \\
    prob-correlation & 0.310 &\textbf{0.027} &\textbf{0.017} &0.476 & &0.232 &0.241 \\
    entropy          & 0.588 &0.075 &\textbf{0.048} &0.763 &0.772 & &0.509 \\
    prob-entropy     & 0.582 &0.071 &\textbf{0.045} &0.745 &0.763 &0.497 & \\
    \hline
    \textbf{slump}
    & standard & random & half & corr. & pr.-corr. & entr. & pr.-entr. \\
    \hline
    standard         & &0.056 &0.129 &\textbf{0.013} &\textbf{0.002} &0.468 &0.421 \\
    random           & 0.946 & &0.630 &0.156 &\textbf{0.053} &0.929 &0.927 \\
    half             & 0.874 &0.375 & &0.114 &\textbf{0.028} &0.851 &0.844 \\
    correlation      & 0.987 &0.848 &0.889 & &0.246 &0.985 &0.984 \\
    prob-correlation & 0.998 &0.949 &0.973 &0.759 & &0.997 &0.997 \\
    entropy          & 0.538 &0.073 &0.152 &\textbf{0.015} &\textbf{0.003} & &0.441 \\
    prob-entropy     & 0.585 &0.075 &0.159 &\textbf{0.017} &\textbf{0.003} &0.565 & \\
    \hline
    \textbf{yacht}
    & standard & random & half & corr. & pr.-corr. & entr. & pr.-entr. \\
    \hline
    standard         & &0.050 &\textbf{0.022} &\textbf{0.013} &\textbf{0.027} &0.468 &0.479 \\
    random           & 0.952 & &0.398 &0.370 &0.381 &0.946 &0.949 \\
    half             & 0.979 &0.608 & &0.421 &0.515 &0.972 &0.972 \\
    correlation      & 0.987 &0.636 &0.585 & &0.650 &0.983 &0.985 \\
    prob-correlation & 0.974 &0.625 &0.491 &0.356 & &0.968 &0.969 \\
    entropy          & 0.538 &0.056 &\textbf{0.029} &\textbf{0.018} &\textbf{0.033} & &0.509 \\
    prob-entropy     & 0.527 &0.053 &\textbf{0.029} &\textbf{0.016} &\textbf{0.032} &0.497 & \\
    \hline
  \end{tabular}
\end{table}

\section{Conclusions}
\label{s5}

Machine learning literature has deeply investigated ensemble models, reporting their advantages with respect to the use of a weak learner. In particular, ensemble models are characterized by different features that could improve the performance of a learning technique: they are able to reduce the variance, the average out biases, they can cover a larger area of the hypothesis space with respect to a single model, and they generally present a better generalization than a single model. Nonetheless, ensemble models were not deeply investigated in the field of GP. This paper answered this call by proposing a strategy to build ensemble models by using genetic programming. Two different GP versions are taken into account: standard syntax-based GP (STGP) and a GP system (GSGP) able to directly include semantic awareness in the search process by using geometric semantic operators.

The main objective of the study was to understand whether an ensemble model made of a blend of GP individuals evolved by both STGP and GSGP can be effectively pruned using different criteria based on correlation and entropy. These criteria are used to avoid the construction of an ensemble where weak learners are semantically too similar. In fact, as existing literature suggested, an ensemble model should be made of accurate and diverse models. The strategies developed were tested over different benchmark problems already considered in the GP literature. Results are interesting and, while they do not allow to draw a general conclusion about the superiority of a criterion with respect to the other ones, they show that considering a similarity criterion when constructing a GP ensemble can help in maintaining the generalization ability of the resulting model while reducing the computational effort.  

This work represents a preliminary study and several future works are planned: first of all, a study aimed at determining the optimal number of weak learners for optimizing the performance of the ensemble over unseen data represent a priority. This would allow practitioners to use the system without the need to determine the number of weak learners, a parameter that has an impact on the performance of the ensemble. Furthermore, the design of pruning criteria that are parameter independent is an important future work to save the user the time required for tuning them. Additionally, it would be interesting to pursue the study of the interaction between syntax and semantics in GP, a very important topic in the field, that is still not well understood.

\section*{Acknowledgements}
This work was also financed through the Regional Operational Programme CENTRO2020 within the scope of the project CENTRO-01-0145-FEDER-000006.

\bibliographystyle{splncs03}
\bibliography{Castelli}

\end{document}